\documentclass[11pt]{article}

\usepackage[preprint]{acl}

\usepackage{times}
\usepackage{latexsym}
\usepackage{adjustbox}
\usepackage{amsmath}
\usepackage{booktabs}
\usepackage{multirow}
\usepackage{enumitem}
\setlist[enumerate]{topsep=0pt, partopsep=0pt, parsep=0pt, itemsep=0pt}
\setlist[itemize]{topsep=0pt, partopsep=0pt, parsep=0pt, itemsep=0pt}

\usepackage[T1]{fontenc}

\usepackage[utf8]{inputenc}

\usepackage{microtype}

\usepackage{inconsolata}

\usepackage{graphicx}

\title{Prototype-Based Dynamic Steering for Large Language Models}

\author{Ceyhun Efe Kayan  \quad Li Zhang \\
  Drexel University \\
  {\tt cek99,harry.zhang@drexel.edu}
}

\begin{document}
\maketitle
\begin{abstract}
Despite impressive breadth, LLMs still rely on explicit reasoning instructions or static, one-fits-all steering methods, leaving a gap for adaptive, instruction-free reasoning amplification. We present Prototype‑Based Dynamic Steering (PDS), a test‑time method that amplifies large language model (LLM) reasoning without adding or altering instructions. We introduce “reasoning prototypes” by clustering activation differences between Chain‑of‑Thought (CoT) and neutral prompts. At inference, an input’s hidden state is projected onto these prototypes to form an instance‑specific steering vector. Evaluated on GSM8K, AQuA‑RAT, and BIG-Bench tasks, PDS consistently improves accuracy without fine‑tuning or prompt engineering. Notably, the gains persist even when CoT is explicitly suppressed to improve cost-efficiency, indicating that the intervention strengthens latent reasoning processes rather than inducing a superficial behavioral shift. These results position dynamic, prototype‑guided steering as a lightweight alternative to training‑time approaches for enhancing LLM reasoning.
\end{abstract}

\section{Introduction}
\label{sec:intro}
The performance of Large Language Models (LLMs) for different tasks across various domains has been remarkable \cite{Brown2020GPT3, Touvron2023LLaMA}. Their ability to generate human-like text, answer questions, and perform complex reasoning tasks has been proven many times. However, their black-box nature poses challenges for safety, reliability, and alignment. A critical research area is the development of methods for steering which is the ability to guide a model's output towards desired attributes (truthfulness, specific tones, jailbreak etc.) or away from undesired ones (toxicity, gender bias etc.) \citep{Wang_2025,konen-etal-2024-style,postmus2025steeringlargelanguagemodels,lamb2025focusthisthatsteering}.

Traditional methods for steering, such as fine-tuning, are computationally expensive, require large datasets and can cause catastrophic forgetting of other desired behaviors \cite{Ouyang2022RLHF, Rafailov2023DPO}. A more recent and lightweight alternative is the use of steering vectors \cite{Turner2023ActivationEngineering, postmus2025steeringlargelanguagemodels}. These vectors, when added to internal activations at specific layers, can effectively shift the model's behavior in a targeted manner. 
With in-context learning, prior work typically constructs these vectors from the difference in activations elicited by carefully chosen pairs with contrast on a specific semantic attribute, such as sentiment, formality, or reasoning presence. These prompts are usually handcrafted or selected from curated datasets to contrast specific behavioral attributes, such as helpfulness, conscientiousness or politeness \citep{konen-etal-2024-style, yang2025exploringpersonalitytraitsllms}.

\begin{figure*}[!tt]
    \includegraphics[width=\textwidth]{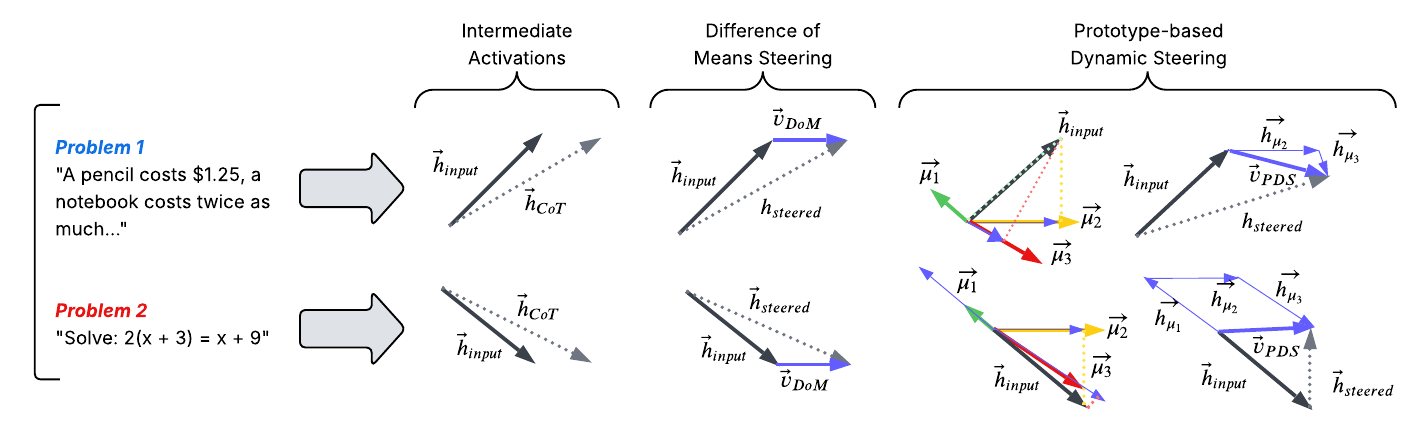}
    \caption{\textbf{Overview of technical differences between Difference-of-Means and proposed Prototype-Based Dynamic Steering}. Instead of using a single vector for each sample like DoM, PDS leverages the projections of input activations onto prototypes to compute the steering vector, which is then injected into the model's residual stream to enhance reasoning behavior.}
    \label{fig:Fig0}
\end{figure*}

The most widely used technique for constructing steering vectors is \textbf{difference-of-means (DoM)}, which defines the steering vector as the difference between the means of the contrastive prompts from the training set. While effective, static techniques like difference-of-means apply the same intervention to every example in a test set regardless of context. A single difference-of-means vector captures an average behavioral shift but cannot adapt to the nuanced requirements of different problems within the same domain. For instance, different mathematical reasoning problems may benefit from different reasoning strategies, yet current methods apply identical steering across all inputs.

In this paper, we propose a robust and data-driven approach for constructing steering vectors during inference time, referred to as \textbf{Prototype-Based Dynamic Steering}. As described in Figure \ref{fig:Fig0} instead of using static steering vectors, we learn a set of ``prototype'' vectors that represent different fine-grained target behaviors from the training set examples. In reasoning tasks specifically, those could be various strategies happening during models' step-by-step reasoning. We discover these prototypes by clustering activation differences between Chain-of-Thought (CoT) prompts \cite{Wei2022ChainOfThought} and neutral prompts, where different clusters may capture distinct reasoning strategies embedded in the activation space. After learning reasoning prototypes, for any new input, we construct a unique, tailored steering vector at inference-time by summing the projections of input activations onto prototypes which allows each prototypes to contribute to the construction of the steering vector according to their match with the ongoing reasoning context. The resulting vector provides a context-aware representation of the ongoing reasoning process at the beginning of the decoding. This way, each test example is guided by its custom steering vector, instead of a one-size-fit-all one in the standard difference-of-means technique.

Our approach is grounded in a fundamentally different view of behavioral steering. Traditional methods such as DoM assume that complex behaviors can be shifted by a single, universal direction in activation space, a “behavioral uniformity” view that considers variation in reasoning approaches as noise. In contrast, we argue that reasoning is not monolithic but arises from combinations of multiple distinct cognitive strategies, each occupying a structured subspace of intermediate activations. PDS is designed to leverage this diversity: by projecting input activations onto a basis of reasoning prototypes, it performs input-responsive steering — dynamically selecting the optimal blend of strategies for the specific problem at hand. This perspective reframes steering as problem-tailored enhancement rather than one-size-fits-all intervention, and it underpins every design choice behind PDS.

To evaluate the effectiveness of our method, we apply Prototype-Based Dynamic Steering across three prompting strategies: three prompting strategies regarding CoT: encouraging (CoT), discouraging (Anti-CoT), and making no special requirement (Neutral). In the Neutral setting, our method successfully induces CoT-like behavior and improves accuracy, demonstrating its ability to enhance reasoning by shifting the model toward step-by-step problem-solving. In the Anti-CoT case, where prompts explicitly discourage step-by-step reasoning, our method does not override such discouragement to evoke CoT, but it still leads to a substantial performance gain, suggesting that steering can also improve reasoning quality without altering the high-level behavior. These findings indicate that our method supports reasoning enhancement through two complementary pathways: (i) behavioral shift toward CoT-style reasoning, and (ii) activation-level refinement that improves performance without extra tokens.

In summary we propose a novel method for discovering behavioral prototypes by clustering activation difference vectors, a dynamic steering mechanism that constructs an input-specific steering vector during inference time by projecting the current activations onto the learned prototypes. We show that PDS outperforms DoM on the \textbf{GSM8k} \cite{Cobbe2021GSM8k}, \textbf{AQuA-RAT} \cite{ling2017program} and \textbf{BIG-Bench} \cite{srivastava2022beyond} benchmarks.



\section{Related Work}
\label{sec:related_work}
A growing body of work \citep{Turner2023ActivationEngineering, Subramani2022ExtractingLatent, Panickssery2024CAA, Li2023ITI, Arditi2024SingleRefusal, Chen2025SEAL, Liu2023InContextVectors, Hernandez2023InspectingEditing, Zou2025OneShot, Lee2025ProgrammingRefusal, Li2025SAEFree} shows that steering vectors enable adjustable control of large language models without altering model weights. We situate our approach within four complementary research directions.

\paragraph{Foundations of Activation Steering}
The linear representation hypothesis holds that semantic features correspond to linear directions in transformer activation spaces \citep{park2025geometrycategoricalhierarchicalconcepts, cunningham2023sparseautoencodershighlyinterpretable}. Early explorations confirm that model behavior, such as sentiment, can be modulated via directed activation interventions \citep{Tigges2023LinearSentiment}. \citet{Subramani2022ExtractingLatent} initiates latent steering by extracting meaningful activation differences, and \citet{Turner2023ActivationEngineering} formalizes activation addition for steering model behavior. \citet{Panickssery2024CAA} advances this via Contrastive Activation Addition (CAA), averaging differences between contrasting data pairs for reliable alignment steering. Recent geometric and norm-preserving techniques such as Householder Pseudo-Rotation \citep{pham-nguyen-2024-householder} and Angular Steering \citep{vu2025angular} address issues of instability in additive steering, offering smoother and more precise behavior control.

\paragraph{Behavioral Steering via Difference-of-Means}
Building on contrastive activation ideas, several studies compute steering vectors by taking differences in mean activations across paired datasets \citep{Turner2023ActivationEngineering, Panickssery2024CAA, Li2023ITI}. For example, \citet{Li2023ITI} uses sparse linear probes to identify truth-oriented directions, achieving substantial performance gains on TruthfulQA \citep{lin-etal-2022-truthfulqa}. Similarly, \citet{Zou2023RepresentationEngineering} extracts activation vectors representing abstract targets like honesty and emotions using this contrastive paradigm. \citet{Arditi2024SingleRefusal} identifies and manipulates a single refusal direction, impacting safety-labeled model behavior. The SEAL method \citep{Chen2025SEAL} further refines CoT output steering by dampening transitional and reflective reasoning steps, improving results on GSM8K and Math500 benchmarks \citep{Cobbe2021GSM8k}.
Moreover, newer contrastive prompting approaches such as Contrastive Prompting \citep{yao2025largelanguagemodelscontrastive} and Learning from Contrastive Prompts\citep{li2024learningcontrastivepromptsautomated} improve reasoning by explicitly eliciting correct and incorrect chains. Spectral Editing of Activations (SEA) \citep{qiu2024spectraleditingactivationslarge} addresses high variance in difference-of-means methods by constructing robust steering directions via covariance-based subspace projection.

\paragraph{Alternative Steering Interventions}
Beyond residual-based activation addition, steering methods also operate on attention activations and across full transformer layers. \citet{Liu2023InContextVectors} demonstrates that attention-layer interventions can steer toxicity and stylistic traits. \citet{Hernandez2023InspectingEditing} explores fine-grained knowledge manipulations to inspect or edit internal representations. Studies like \citet{Zou2025OneShot} directly optimize steering vectors using single examples with gradient-based methods, leading to high success in suppressing unwarranted refusal behavior across safety benchmarks like HarmBench \citep{mazeika2024harmbenchstandardizedevaluationframework}. \citet{Lee2025ProgrammingRefusal} treat each transformer's layer bias term as a trainable vector, leveraging this structure to match or exceed fine-tuned refusal control without weight updates. 

\paragraph{Prototype-Based \& Sparse Autoencoder Methods}
Prototype-driven frameworks add interpretability and causal clarity to steering. Sparse autoencoder (SAE)-based methods also extract feature directions and monosemantic latent features from raw CoT traces or residual activations, enhancing reasoning by overcoming polysemanticity and aiding precise behavioral intervention through more interpretable latent features \citep{Li2025SAEFree, cunningham2023sparseautoencodershighlyinterpretable}. These approaches help pinpoint which latent features directly cause behavior and can be manipulated for targeted control. Studies further consolidates sparse-autoencoder techniques for interpreting and steering LLMs \citep{shu2025surveysparseautoencodersinterpreting}.
Prototype-based interpretability has shown promise beyond vision applications: NLP extensions of prototypical networks demonstrates that models could learn inherently interpretable embeddings during fine-tuning, capturing representative features for downstream tasks while maintaining competitive performance and transparency at both token and sample levels \citep{xie-etal-2023-proto}.
Mechanistic and geometric analyses reveal structured, often orthogonal subspaces in LLM representations, underpinning many steering techniques that treat high-level behaviors as latent directions \citep{geva-etal-2021-transformer,park2025geometrycategoricalhierarchicalconcepts}.

\paragraph{Positioning of Our Work}
We build on these insights but depart in two key ways. First, instead of a single difference-of-means vector, we learn a \emph{prototype set}: a small basis that captures diverse facets of the target behavior, making the intervention more interpretable and robust. Second, by projecting inference-time activations onto learned prototypes, the steering vector for the input is constructed as a function of the information related to the problem at hand. Collectively, the literature demonstrates that steering vectors are an emerging technique for interpretable LLM alignment. Our contribution extends this paradigm by introducing prototype-based steering.

\begin{figure}[!t]
\centering
    \includegraphics[width=0.35\textwidth]{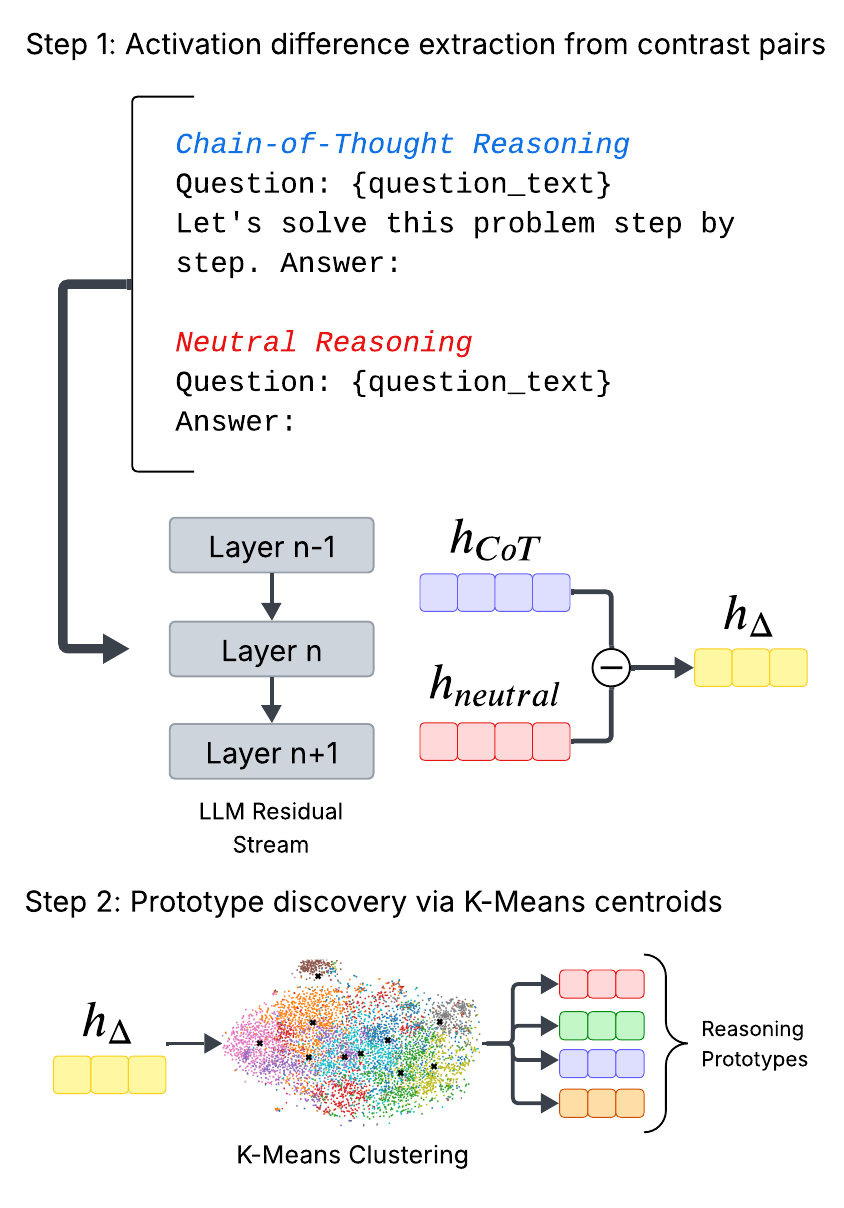}
    \caption{\textbf{Overview of prototype discovery in Prototype-Based Dynamic Steering}. Activation differences are extracted for the contrast input pairs. Centroids obtained after K-Means clustering are treated as reasoning prototypes.}
    \label{fig:Fig1}
\end{figure}

\section{Methods}
\label{sec:method}
Our method consists of three stages: (1) collecting activation difference vectors representing CoT inducing behavior, (2) discovering a set of ``reasoning prototypes'' by clustering these difference vectors, and (3) using these prototypes to construct an input-specific steering vector for each input dynamically, during inference. We provide theoretical justification for each component and detailed analysis of the geometric properties of the resulting prototype space.
\paragraph{Activation Difference Collection}
The foundation of our approach rests on the hypothesis that behavioral transformations in neural networks can be captured as directions in activation space \citep{park2025geometrycategoricalhierarchicalconcepts, cunningham2023sparseautoencodershighlyinterpretable}. For reasoning, we seek to identify the transformation from neutral problem-solving to explicit step-by-step reasoning while preserving the nuanced variations that different problem types require.
For each training example $x_i$ from a particular dataset's training split, we generate two carefully designed prompt variants that isolate the reasoning transformation:
\begin{itemize}
    \item $p_{i, CoT}$: The CoT prompt, e.g.,``Question: [problem text]... Let's think step by step. Answer: ''
    \item $p_{i, neutral}$: The neutral prompt, e.g., ``Question: [problem text]... Answer:''
\end{itemize}
The CoT prompt explicitly encourages step-by-step reasoning through the instruction phrase, while the neutral prompt represents the model's default problem-solving approach without explicit reasoning guidance. This design ensures that activation differences can be attributed specifically to the reasoning transformation rather than confounding task-specific variations.
We process both prompts through the model and extract the final token's hidden state activation $h_l$ at layer $l$. We empirically determine that layer 16 (out of 32 total layers) provides optimal signal for reasoning behaviors, consistent with prior work suggesting middle layers capture higher-level semantic processing \citep{liu2023lostmiddlelanguagemodels,skean2024doesrepresentationmatterexploring,zheng2025attention}. From these activations, we compute a set of $N$ activation difference vectors:
$$D = \{d_1, d_2, ..., d_N\}$$
$$d_i = h_{l}(p_{i, CoT}) - h_{l}(p_{i, neutral})$$
Each vector $d_i$ represents a specific guidance for reasoning transformation for problem $x_i$. Crucially, these vectors does not simply capture the \textbf{average} shift toward reasoning like DoM, but encode \textbf{each specific} problem’s idiosyncratic characteristics that may include solution strategies.

\paragraph{Prototype Discovery via Clustering}
Traditional difference-of-means approach computes the average activation difference vector, effectively discarding the rich diversity present in the set of activation differences. Our central insight is that this diversity contains valuable information about distinct reasoning strategies that should be preserved and leveraged rather than averaged away.
We apply k-means clustering \citep{MacQueen1967,Lloyd1982} to discover latent structure in the difference vector space:
$$\{\mu_1, \mu_2, ..., \mu_k\} = \text{k-means}(D, k)$$
The choice of k-means clustering is motivated by several theoretical and practical considerations. K-means groups activation differences into clusters that correspond to distinct problem solving strategies. The centroid of each cluster represents an average activation shift towards the associated reasoning style, making it usable as a targeted steering direction.
The number of clusters $k$ is determined via the elbow method, using the within-cluster sum of squares criterion.
Each cluster centroid $\mu_j$ represents a reasoning prototype - a canonical direction in activation space corresponding to a specific reasoning strategy. 
\paragraph{Theoretical Justification for Prototype-Based Steering}
We provide theoretical motivation from complementary perspectives to answer the question ``Why should prototype-based clustering capture meaningful behavioral structure?''
\begin{itemize}
    \item[1.]\textbf{Geometric Perspective:} If reasoning behaviors occupy a structured subspace in activation space, then clustering activation differences should recover the principal directions of this subspace. The success of our method provides empirical evidence that mathematical reasoning indeed occupies such a structured region, validating the linear representation hypothesis for complex cognitive behaviors. \citep{park2025geometrycategoricalhierarchicalconcepts, cunningham2023sparseautoencodershighlyinterpretable}.
    \item[2.]\textbf{Cognitive Perspective:} Mathematical problem-solving involves multiple strategies that share core computational principles but differ in execution details. Clustering activations identifies these strategic variations, enabling dynamic selection of appropriate reasoning approaches based on problem-specific characteristics.
    \item[3.]\textbf{Information Theoretic Perspective:} Rather than collapsing the information in activation differences to a mean vector, clustering preserves diversity while reducing dimensionality from $N$ individual examples to $k$ interpretable prototypes, balancing information retention with computational efficiency.
\end{itemize}
\begin{table}[!t]
\centering
\resizebox{\columnwidth}{!}{%
\begin{tabular}{lcccccc}
\toprule
\multirow{2}{*}{\textbf{Condition}} & \multicolumn{3}{c}{\textbf{GSM8k}} & \multicolumn{3}{c}{\textbf{Aqua-Rat}} \\
\cmidrule(lr){2-4} \cmidrule(lr){5-7}
& \textbf{Baseline} & \textbf{DoM} & \textbf{PDS} & \textbf{Baseline} & \textbf{DoM} & \textbf{PDS} \\
\midrule
CoT        & 71\% & 75\% & \textbf{78\%} & 49\% & 50\% & \textbf{53\%} \\
Neutral    & 68\% & 74\% & \textbf{78\%} & 46\% & 50\% & \textbf{53\%} \\
Anti-CoT   & 11\% & 18\% & \textbf{21\%} & 29\% & 30\% & \textbf{31\%} \\
\bottomrule
\end{tabular}
}
\caption{Performance comparison (Accuracy \%) across prompting conditions: Baseline (no steering), Difference of Means (DoM), and Prototype-Based Dynamic Steering (PDS) on GSM8k and AQuA-Rat benchmarks.}
\label{tab:dom_comparison}
\end{table}
\paragraph{Dynamic Steering via Projection}
Our steering mechanism is fundamentally dynamic and input-adaptive. For inference on a new problem, we first pass the input prompt through the model (without generation) to obtain activations $h_{input}$ of the final input token at layer $l$. This vector encodes the model's initial ``understanding'' of the problem before generation begins.
We then construct a custom steering vector $v_{steer}$ by projecting $h_{input}$ onto our learned prototype set:
$$v_{steer}(h_{input}) = \sum_{j=1}^{k} \text{proj}_{\mu_j}(h_{input})$$
This projection operation exhibits 3 crucial properties:
\begin{itemize}
    \item \textbf{Adaptive Weighting:} The magnitude of the constructed steering vector is computed based on the alignment between each prototype and the current input's activation pattern. Problems naturally activating reasoning-related representations produce larger weights for relevant prototypes, inducing an adaptive weighting of the relevant prototype.
    \item \textbf{Subspace Preservation:} The sum of projections yields a vector lying entirely within the subspace spanned by the prototypes, ensuring steering remains within the reasoning manifold rather than venturing into semantically irrelevant directions.
    \item \textbf{Compositional Strategy Selection:} Each projection weight indicates the contribution of a specific reasoning strategy to the final steering vector, enabling interpretable analysis of which cognitive approaches the model considers relevant for each problem.
\end{itemize}
The steering intervention modifies activations via:
$$h'_{input} = h_{input} + \alpha \cdot v_{steer}(h_{input})$$
where $\alpha$ is a hyperparameter controlling steering strength. We apply this intervention selectively to the first output token only, maximizing impact on reasoning initiation while minimizing interference with subsequent generation dynamics.

\begin{figure}[ht]
\centering
    \includegraphics[width=0.35\textwidth]{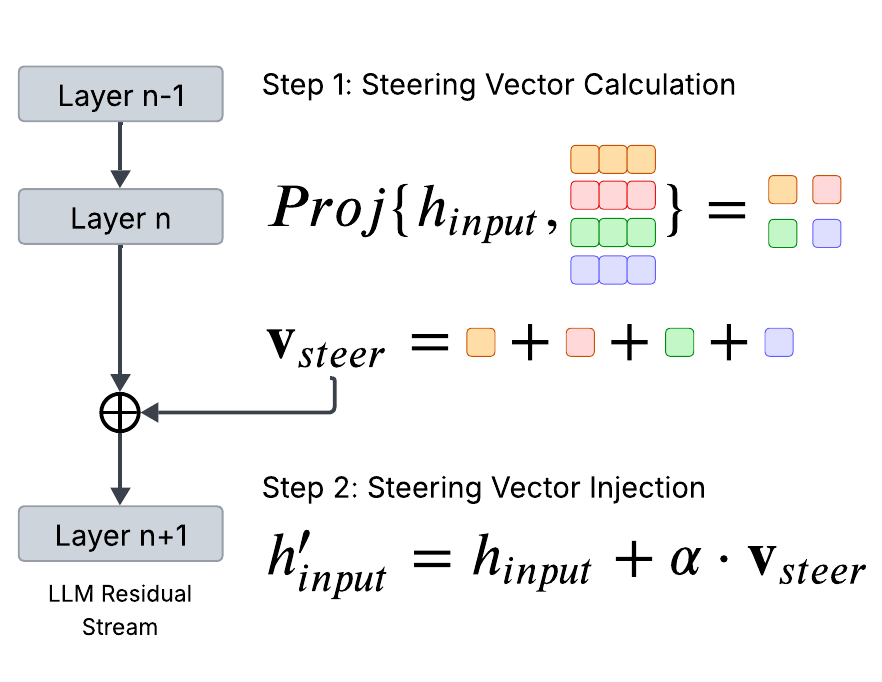}
    \caption{\textbf{Overview of steering vector injection in Prototype-Based Dynamic Steering}. During inference, projections of input activations onto prototypes are summed to compute the steering vector, which is then injected into the model's residual stream to enhance reasoning behavior.}
    \label{fig:Fig2}
\end{figure}
\paragraph{Geometric Analysis of the Prototype Space}
Our empirical analysis reveals remarkable geometric properties of the learned prototypes. Computing pairwise angles between prototypes yields a narrow geometric composition where angular separations are consistently within the narrow range of 8-22 degrees across different datasets and experimental conditions.
This tight angular clustering indicates that all prototypes lie approximately on the surface of a high-dimensional cone in activation space.\\
\textbf{The Cone's Central Axis:} The central axis represents the core cognitive transformation from ``direct answering'' to ``step-by-step reasoning''. Mathematically, this axis corresponds to $\sum_{j=1}^{k} \mu_j = k \cdot \text{mean}(D)$, equivalent to the traditional difference-of-means vector scaled by the number of clusters. This reveals that our prototype-based approach contains difference-of-means as a special case while extending significantly beyond it.\\
\textbf{The Cone's Surface:} Each prototype captures a strategic variation around the core reasoning transformation. These variations correspond to different problem-solving styles, domain-specific approaches (arithmetic vs. algebraic vs. geometric), and different levels of mathematical abstraction. PDS can capture these variances, unlike difference-of-means which discards it by averaging. PDS can also amplify the representation of input-specific information, since the dynamic projection selects optimal strategy combinations based on problem activation patterns.

\begin{table*}[!t]
\small
\centering
\begin{adjustbox}{width=\linewidth}
\begin{tabular}{lccc|ccc|ccc}
\toprule
\multirow{2}{*}{\textbf{Dataset}} & \multicolumn{3}{c|}{\textbf{Chain-of-Thought}} & \multicolumn{3}{c|}{\textbf{Neutral}} & \multicolumn{3}{c}{\textbf{Anti-Chain-of-Thought}} \\
\cmidrule(lr){2-4} \cmidrule(lr){5-7} \cmidrule(lr){8-10}
& \textbf{No-Steering} & \textbf{DoM} & \textbf{PDS} & \textbf{No-Steering} & \textbf{DoM} & \textbf{PDS} & \textbf{No-Steering} & \textbf{DoM} & \textbf{PDS} \\
\midrule
Checkmate-in-One     & 0.2\% & \textbf{0.6\%} & \textbf{0.6\%} & \textbf{0.2\%} & \textbf{0.2\%} & 0.0\% & \textbf{0.2\%} & 0.0\% & \textbf{0.2\%} \\
Physics Questions     & 0.0\% & 0.0\% & 0.0\% & 0.0\% & 0.0\% & 0.0\% & 0.0\% & 0.0\% & 0.0\% \\
Rhyming               & 23.7\% & \textbf{25.8\%} & 24.7\% & 23.7\% & \textbf{25.8\%} & 23.7\% & 24.2\% & \textbf{24.7\%} & \textbf{24.7\%} \\
Temporal Sequences    & 59.5\% & 50.5\% & \textbf{60.0\%} & 57.0\% & 50.5\% & \textbf{58.5\%} & \textbf{60.0\%} & 50.5\% & 59.5\% \\
Intersect Geometry    & 34.6\% & \textbf{38.8\%} & 35.4\% & 34.3\% & 36.6\% & \textbf{41.8\%} & 31.2\% & \textbf{36.2\%} & 32.2\% \\
\bottomrule
\end{tabular}
\end{adjustbox}
\caption{Performance comparison (Accuracy \%) across prompting conditions: No Steering, Difference of Means (DoM), and Prototype-Based Dynamic Steering (PDS) on BIG-Bench benchmark with Llama-3-Instruct-8B model.}
\label{tab:dom_comparison_8B}
\end{table*}

\begin{table*}[!t]
\centering
\small
\begin{adjustbox}{width=\linewidth}
\begin{tabular}{lccc|ccc|ccc}
\toprule
\multirow{2}{*}{\textbf{Dataset}} & \multicolumn{3}{c|}{\textbf{Chain-of-Thought}} & \multicolumn{3}{c|}{\textbf{Neutral}} & \multicolumn{3}{c}{\textbf{Anti-Chain-of-Thought}} \\
\cmidrule(lr){2-4} \cmidrule(lr){5-7} \cmidrule(lr){8-10}
& \textbf{No-Steering} & \textbf{DoM} & \textbf{PDS} & \textbf{No-Steering} & \textbf{DoM} & \textbf{PDS} & \textbf{No-Steering} & \textbf{DoM} & \textbf{PDS} \\
\midrule
Checkmate-in-One     & 9.8\% & 6.8\% & \textbf{10.4\%} & 7.0\% & 7.8\% & \textbf{7.8\%} & 14.8\% & 10.8\% & \textbf{15.8\%} \\
Physics Questions     & 0.0\% & 0.0\% & 0.0\% & 0.0\% & 0.0\% & 0.0\% & 0.0\% & \textbf{6.2\%} & \textbf{6.2\%} \\
Rhyming               & 23.7\% & 14.2\% & \textbf{25.8\%} & \textbf{22.6\%} & 15.8\% & \textbf{22.6\%} & \textbf{26.8\%} & 25.3\% & 25.8\% \\
Temporal Sequences    & \textbf{98.5\%} & 97.0\% & \textbf{98.5\%} & \textbf{98.5\%} & 97.5\% & \textbf{98.5\%} & \textbf{98.5\%} & 97.5\% & \textbf{98.5\%} \\
Intersect Geometry    & \textbf{38.4\%} & 37.4\% & \textbf{38.4\%} & 37.6\% & \textbf{40.0\%} & 37.2\% & \textbf{41.4\%} & 40.2\% & 40.6\% \\
\bottomrule
\end{tabular}
\end{adjustbox}
\caption{Performance comparison (Accuracy \%) across prompting conditions: No Steering, Difference of Means (DoM), and Prototype-Based Dynamic Steering (PDS) on BIG-Bench benchmark with Llama-3-Instruct-70B model.}
\label{tab:dom_comparison_70B}
\end{table*}

\section{Evaluation}
\label{sec:results}

Our evaluation spans three common reasoning benchmarks. For each dataset, we (1) collect CoT-neutral activation deltas on the training split, (2) cluster them into reasoning prototypes, and (3) evaluate under Neutral, CoT, and Anti-CoT prompt conditions using Accuracy@1.
\begin{itemize}
    \item \textbf{GSM8K:} Grade-school math word problems requiring multi-step arithmetic reasoning. \citep{Cobbe2021TrainingVT}
    \item \textbf{AQuA-RAT:} Multiple-choice algebraic word problems that demand symbolic manipulation and higher-order reasoning. \citep{ling2017program}
    \item \textbf{BIG-Bench Subset:} We selected five tasks that exercise distinct skills: checkmate-in-one, physics-questions, rhyming, temporal sequences, intersect geometry. \citep{srivastava2022beyond}
\end{itemize}

We conduct our experiments using the LLaMA-3-Instruct-8B \citep{Touvron2023LLaMA} language model. To account for the increased difficulty of the selected BIG-Bench subset, where tasks such as Checkmate-in-One and Physics Questions present significant challenges for smaller models, we additionally evaluate performance using the larger LLaMA-3-Instruct-70B model. This comparison allows us to assess whether our prototype-based steering method scales effectively with model capacity and to verify that observed improvements are not limited by the expressiveness of the 8B model. Evaluating on both 8B and 70B variants enables a more comprehensive understanding of steering efficacy across model sizes.

Each dataset in our evaluation suite demands diverse and heterogeneous reasoning strategies, even within seemingly domains. While GSM8K and AQuA-RAT are both math-focused, the problems span a variety of reasoning types. For example, GSM8K questions require combining numerical estimation, unit conversion, logical sequencing, or multi-step arithmetic decomposition (e.g., ``If a pencil costs \$1.25 and a notebook costs twice as much, how much do five notebooks and three pencils cost in total?''). Similarly, AQuA-RAT tasks involve symbolic manipulation, pattern recognition in algebraic expressions, distractor elimination in multiple-choice formats, and even linguistic disambiguation in problem phrasing (e.g., ``What is the value of $x$ if three times the difference between $x$ and $4$ is equal to $18$?''). These subtasks demand distinct cognitive strategies within a single domain. The BIG-Bench subset expands this diversity further by explicitly testing spatial, temporal, linguistic, and logic-based reasoning, such as determining move sequences in chess (Checkmate-in-One), resolving event orders (Temporal Sequences), or rhyming constraints in generative tasks.
\begin{itemize}
    \item \textbf{Neutral:} Provides no reasoning cues. \textbf{Question:} “[problem statement]”. \textbf{Answer:}
    \item \textbf{Chain-of-Thought (CoT):} Encourages the model to reason step by step. \textbf{Question:} “[problem statement]”. Let’s think step by step. \textbf{Answer:}
    \item \textbf{Anti Chain-of-Thought (Anti-CoT):} Explicitly instructs the model to avoid reasoning. \textbf{Question:} “[problem statement]”. Answer immediately, without elaboration. \textbf{Answer:}
\end{itemize}
Our prototype-based method is applied separately to each dataset; steering vectors are extracted from that dataset’s training set, clustered into reasoning prototypes, and used for evaluation solely on that dataset. This design avoids any unintended cross-task leakage and ensures that prototype representations are aligned with the unique reasoning requirements of each benchmark.
We evaluate three different experiment cases:
\begin{itemize}
    \item \textbf{No Steering:} The model receives no intervention at the activation level.
    \item \textbf{Difference of Means (DoM):} A global steering direction is computed as the mean activation difference between CoT and Neutral responses, then applied to each new sample for steering.
    \item \textbf{Prototype-Based Dynamic Steering (PDS):} Activation differences are clustered into semantically meaningful prototypes, and the most relevant one is dynamically selected at inference time based on projection alignment with the model’s hidden state.
\end{itemize}

This evaluation structure allows us to test whether fine-grained, interpretable, and dynamically applied steering vectors can enhance model performance across varied reasoning demands, both within and across domains.

\section{Results \& Discussion}
\label{sec:discussion}

\paragraph{Results Across Datasets}

Tables \ref{tab:dom_comparison},\ref{tab:dom_comparison_8B} and \ref{tab:dom_comparison_70B} present our experimental results alongside the difference-of-means (DOM) technique in all three benchmarks. Our prototype-based dynamic steering method demonstrates consistent improvements across all datasets and prompting conditions against the baseline. The results demonstrate remarkable consistent improvements across all three benchmarks. Our dynamic steering method achieves improvements ranging from 0.6\% to 9.66\% across different conditions and benchmarks. The most significant gains consistently occur in the Neutral and Anti-CoT conditions, where the model lacks explicit reasoning guidance. Notably, even when CoT prompting is already present, our method does not cause performance degradation, suggesting that activation-level steering modify reasoning patterns that prompting already has elicited.

\paragraph{Comparison with Difference of Means}

Our method outperforms Difference of Means technique across various conditions and datasets in terms of performance. While both methods achieve improvements over the unsteered baseline, the superior performance of our approach validates the importance of capturing behavioral diversity rather than averaging it away. To investigate whether our improvements stem from directional guidance or simple activation scaling, we conduct an experiment using normalized prototypes. Our original prototypes had an average norm of 4.51, representing significant magnitude in the activation space. When the prototypes are normalized to unit length, performance remains unchanged across all conditions and datasets. This result provides evidence that our projection-based steering method operates through directional guidance rather than magnitude scaling.


Our results demonstrate that PDS achieves consistent improvements up to 9.72\% across three diverse benchmarks. Consistent improvements across GSM8k, AQuA-Rat, and BIG-Bench suggests that PDS enhances the capturing of various patterns in one steering vector. The largest gains consistently occur in Neutral and Anti-CoT settings, indicating PDS is most effective when explicit reasoning guidance is absent, sufficiently compensating for weak reasoning tendencies.

\paragraph{Effect on Completion Length}

We analyzed the average number of tokens generated on the AQuA-RAT dataset. While all three methods receive the same input prompt, the steering interventions influence not only accuracy but also the length and structure of the completions.

\begin{table}[t!]
\centering
\small
\begin{tabular}{cccc}
\toprule
\textbf{Prompt Type} & \textbf{Method} & \textbf{Avg. Tokens} & \textbf{Std. Dev.} \\
\midrule
\multirow{3}{*}{Neutral} 
    & No Steering & 287.31 & 492.82 \\
    & DoM         & 256.96 & 358.54 \\
    & PDS         & \textbf{254.58} & \textbf{357.60} \\
\cmidrule{1-4}
\multirow{3}{*}{CoT}     
    & No Steering & 275.61 & 358.16 \\
    & DoM         & 303.09 & 440.70 \\
    & PDS         & \textbf{258.03} & \textbf{146.45} \\
\bottomrule
\end{tabular}
\caption{\textbf{Average number of output tokens on the AQuA-RAT dataset across prompting types and steering methods}. CoT prompts produce longer, Anti-CoT prompts yield concise completions. PDS improves accuracy in all settings with minimal increase in length.}
\label{tab:token_length_aqua}
\end{table}

PDS completions are slightly longer on average, suggesting more structured or elaborated reasoning chains. Notably, the increased length is accompanied by a substantial gain in accuracy, indicating that the additional tokens reflect meaningful intermediate reasoning rather than verbosity.

\paragraph{Reasoning Enhancement without CoT}

The most impactful finding of our study is the model’s behavior under Anti-Chain-of-Thought (Anti-CoT) prompts, which explicitly instruct the model to “answer immediately without elaboration.” Conventional activation-steering methods often override such instructions, altering output structure. In contrast, PDS preserves the requested response format while still improving problem-solving capability.

Across benchmarks, PDS consistently boosts accuracy in Anti-CoT settings without inducing step-by-step reasoning. As shown in Table 1, GSM8K accuracy rises from 11\% (no steering) to 21\% (+10 \%), and AQuA-RAT improves from 29\% to 31\% (+2 \%). Similarly, in the BIG-Bench tasks (Table 5), PDS achieves +6.2 pp on Physics Questions and a modest +1\% on Checkmate-in-One, while preserving the terse output style mandated by Anti-CoT instructions.
This outcome is a core finding of our work and the main proof of PDS's contribution to enhancing latent reasoning capabilities. Unlike prior steering approaches that primarily seek to change model behavior, PDS demonstrates that substantial reasoning gains can be achieved without altering instruction-level outputs. By selectively strengthening reasoning while preserving compliance with instructions, PDS establishes a robust intervention paradigm for solving problems that require diverse reasoning approaches.

\begin{table}[t!]
\centering
\small
\begin{tabular}{lccc}
\toprule
\textbf{Dataset} & \textbf{No Steering} & \textbf{PDS} & \textbf{$\Delta$} \\
\midrule
GSM8k    & 11.60\% & 21.12\% & +9.52\% \\
Aqua-Rat & 29.50\% & 31.50\% & +2.00\% \\
\midrule
Checkmate-in-One  & 14.8\% & 15.8\% & +1\% \\
Physics Questions & 0\% & 6.2\% & +6.2\% \\
Rhyming & 26.8\% & 25.8\% & -1\% \\
Temporal Sequences & 98.5\% & 95.8\% & +0\% \\
Intersect Geometry & 41.4\% & 40.6\% & -0.8\% \\
\bottomrule
\end{tabular}
\caption{Performance on Anti-CoT prompting}
\label{tab:anticot_results}
\end{table}


\section{Conclusion}
\label{sec:conclusion}
Prototype-Based Dynamic Steering (PDS) is an inference-time method that enhances reasoning in large language models without altering weights or prompts. By clustering activation differences between CoT and neutral prompts, PDS uncovers interpretable prototypes representing diverse reasoning strategies. Input activations are projected onto these prototypes to generate context-sensitive steering vectors, outperforming the standard difference-of-means approach across benchmarks and prompting styles. Crucially, PDS boosts performance even when explicit step-by-step instructions are omitted, revealing latent reasoning structure in activation space. Future work may extend prototype discovery techniques and apply dynamic steering cross-modally.

\section{Limitations}
While Prototype-Based Dynamic Steering (PDS) demonstrates strong reasoning improvements and introduces a novel activation-space perspective, it has several limitations. First, our current prototype discovery process relies on a predefined dataset of reasoning traces represented as activation differences and may not generalize optimally to domains with different reasoning styles. Second, the effectiveness is not uniform across all tasks according to the empirical outcomes. In particular, improvements tend to be more modest in settings where the model’s baseline reasoning capability is already near saturation, leaving less latent structure to exploit. Third, tasks that rely more heavily on domain-specific knowledge rather than on compositional or step-by-step reasoning offer fewer opportunities for PDS to induce substantial gains. These observations suggest that the strength of PDS is closely tied to the presence of underutilized reasoning capacity in the base model, which is an insight that points to future directions in designing steering strategies tailored to different reasoning regimes. Finally, as with other activation-level interventions, steering techniques can be misused to create jailbreak cases or amplify undesired behaviors. While our work focuses on reasoning enhancement, the fact that PDS provides an alternative mechanism for constructing steering vectors highlights the importance of developing complementary safety measures to mitigate potential jailbreak risks.

\bibliography{custom}

\appendix
\section{Appendix}
\section*{Dataset Details}

We evaluate our method across three complementary reasoning benchmarks, selected for their diversity in problem format, reasoning complexity, and domain coverage. For each dataset, we use a subset of the training split to extract steering prototypes and evaluate performance on the official test set.

\begin{itemize}
    \item \textbf{GSM8K:} GSM8K consists of grade-school math word problems requiring multi-step arithmetic reasoning. We use all of the 7470 samples from the training set to compute steering vectors and evaluate on the full test set of 1,319 examples.
    \item \textbf{AQuA-RAT:} AQuA-RAT is a dataset of multiple-choice algebraic word problems that often involve symbolic manipulation and intermediate computation steps. We sample 10,000 questions from the training set to derive steering representations and test on a 254-example evaluation set.
    \item \textbf{BIG-Bench Subset} From the BIG-Bench benchmark, we select five tasks that reflect distinct reasoning types: \textit{Checkmate-in-One}, \textit{Physics Questions}, \textit{Rhyming}, \textit{Temporal Sequences}, and \textit{Intersect Geometry}. Table REF presents the number of samples taken from each set (training, test) for each task.
\end{itemize}

\begin{table}[ht]
\small
\centering
\begin{tabular}{lcc}
\toprule
\textbf{Dataset} & \textbf{Training Set} & \textbf{Test Set} \\
\midrule
Checkmate-in-One     & 1000 & 699 \\
Physics Questions    & 38 & 16 \\
Rhyming               & 763 & 190 \\
Temporal Sequences    & 800 & 200 \\
Intersect Geometry    & 1000 & 1000\\
\bottomrule
\end{tabular}
\caption{Number of samples used for prototype generation and evaluation for BIG-Bench dataset.}
\label{tab:BIGbenchsamples}
\end{table}

\section*{Prototype Cluster Descriptions}

To interpret the latent structure captured by Prototype-Based Dynamic Steering (PDS), we collect 50 representative prompts from each cluster and query multiple large language models (GPT-4.5, Claude Sonnet 4, Gemini 2.5 Pro) to assign semantic labels. Each model independently analyzed 50 randomly sampled questions from each prototype's cluster to assign a cluster name and explain each cluster without any context and information about our methodology. Cluster-level names for each model and dataset are provided in Tables \ref{tab:gsm8kprompts}, \ref{tab:aquaratprompts} and \ref{tab:BBprompts}.

\paragraph{Philosophy of Prototype-Based Steering}

Our approach embodies a fundamentally different philosophical stance toward behavioral steering compared to traditional methods.

Traditional steering methods, particularly Difference of Means, operate under the assumption that complex behaviors can be captured by single, universal directions in the activation space. This ``behavioral uniformity'' hypothesis treats variations in reasoning approaches as noise to be averaged away. Our prototype-based philosophy challenges this assumption by recognizing that \textbf{complex cognitive behaviors} such as mathematical reasoning are formed by unique combinations of multiple distinct ``reasoning strategies''. Rather than viewing this diversity as problematic variation, we treat it as valuable information that should be preserved and leveraged.

Our philosophy extends beyond static intervention to embrace \textbf{input-responsive steering}. The projection mechanism dynamically selects the optimal combination of reasoning strategies based on the current problem's activation pattern. This represents a shift from ``one-size-fits-all'' steering to ``problem-tailored'' enhancement. This approach reflects a deeper understanding of how cognitive behaviors manifest in neural networks: rather than being represented by single directions, complex behaviors occupy structured vector subspaces that require nuanced intervention strategies.

\begin{table*}[ht]
\centering
\small
\begin{tabular}{@{}c@{}c@{}c@{}c} 
\toprule
\textbf{Cluster} & \textbf{GPT-4.5} & \textbf{Gemini 2.5 Pro} & \textbf{Claude Sonnet-4}\\
\midrule
\textbf{0} & Multi-step Arithmetic Problems & Multi-step Arithmetic Problems & Multi-Step Calculations \\
\textbf{1} & Comparative Quantity Problems & Relational \& Dependent Variables & Complex Multi-Variable Problems\\
\textbf{2} & Basic Quantitative Reasoning & Straightforward Two-Step Calculations & Basic Arithmetic Applications \\
\textbf{3} & Complex Multi-condition Problems & Complex Proportional Reasoning & Business and Real-World Scenarios\\
\textbf{4} & Financial Arithmetic Problems & Rate, Cost, and Profit & Cost Analysis
\\
\textbf{5} & Ratio and Comparative Age Problems & Logical Deduction Puzzles & Fraction and Percentage Problems
\\
\textbf{6} & Arithmetic with Units \& Conversions & Comparative Relational Arithmetic & Simple Relational Problems \\
\textbf{7} & Sequential Operations \& Patterns & Complex Narrative Problems & Complex Logical Reasoning\\
\textbf{8} & Percentage and Fraction Problems & Direct Rate and Unit Problems & Direct Calculation Problems\\
\bottomrule
\end{tabular}
\caption{Cluster interpretability validation for GSM8k dataset via blind evaluation.}
\label{tab:gsm8kprompts}
\end{table*}

\begin{table*}[ht]
\centering
\small
\begin{tabular}{@{}c@{}c@{}c@{}c} 
\toprule
\textbf{Cluster} & \textbf{GPT-4.5} & \textbf{Gemini 2.5 Pro} & \textbf{Claude Sonnet-4}\\
\midrule
\textbf{0} & Rhyming Choices & Phonetic Sound Recognition & Rhyming Word Recognition \\
\textbf{1} & Shape Intersections & Intersection Analysis & Geometric Intersection Counting
\\
\textbf{2} & Chess Mate-One & Chess Checkmate Puzzles & Chess Mate-in-One Puzzles \\
\textbf{3} & Time Interval Deduction & Logical Word Problems & Schedule Logic Reasoning\\

\bottomrule
\end{tabular}
\caption{Cluster interpretability validation for BIG-Bench dataset via blind evaluation.}
\label{tab:aquaratprompts}
\end{table*}

\begin{table*}[ht]
\centering
\small
\begin{tabular}{@{}c@{}c@{}c@{}c} 
\toprule
\textbf{Cluster} & \textbf{GPT-4.5} & \textbf{Gemini 2.5 Pro} & \textbf{Claude Sonnet-4}\\
\midrule
\textbf{0} & Diverse Multi-Step Word Problems & Advanced Word Problems & Complex Multi-Step Problems \\
\textbf{1} & Basic Numerical Computation & Number Properties and Operations & Basic Mathematical Operations\\
\textbf{2} & Financial \& Work-Rate Scenarios & Financial and Proportional Reasoning & Business and Investment Problems \\
\textbf{3} & Fundamental Arithmetic Concepts & Foundational Math Concepts & Elementary Calculations\\
\textbf{4} & Geometry \& Probability Applications & Complex Quantitative Scenarios & Applied Mathematical Reasoning
\\
\textbf{5} & Intensive Numeric Challenges & Basic Arithmetic and Number Sense & Computational Mathematic
\\
\textbf{6} & Complex Algebraic \& Rate Problems & Rate and Change Problems & Intermediate Problem Solving \\
\textbf{7} & SApplied Real-World Quant Problems & Applied Business and Finance & Advanced Word Problems\\
\bottomrule
\end{tabular}
\caption{Cluster interpretability validation for AQuA-RAT dataset via blind evaluation.}
\label{tab:BBprompts}
\end{table*}

\section*{Inference and Steering Hyperparameters}

All experiments use standard autoregressive decoding with temperature $0.1$ and top-$p$ sampling at $0.9$ during prototype collection. Evaluation is performed using greedy decoding with a maximum length of 512 tokens. We fix the intervention layer to 16 for all steering experiments, based on preliminary probing of reasoning signal strength across layers.

The steering coefficient $\alpha$ is held constant at 1.0 across datasets and prompt types. This consistent configuration allows us to isolate the effect of steering vector construction (PDS vs. DoM) without confounding from hyperparameter tuning.

\begin{table*}[ht]
\centering
\small
\begin{tabular}{lc}
\toprule
\textbf{Parameter} & \textbf{Value} \\
\midrule
Model & LLaMA-3-Instruct-8B \\
Temperature & 0.7 \\
Top-$p$ (nucleus sampling) & 0.9 \\
Max tokens & 4096 \\
Sampling strategy & Temperature Sampling \\
\bottomrule
\end{tabular}
\caption{Inference hyper-parameters used across all evaluations.}
\label{tab:inference_hparams}
\end{table*}

\begin{table*}[ht]
\centering
\small
\begin{tabular}{lccc}
\toprule
\textbf{Dataset} & \textbf{Prompt Type} & \textbf{Intervention Layer} & \textbf{Steering Strength ($\alpha$)} \\
\midrule
\multirow{3}{*}{GSM8K} 
    & Neutral    & 16 & 1 \\
    & CoT        & 16 & 1 \\
    & Anti-CoT   & 16 & 10 \\
\midrule
\multirow{3}{*}{AQuA-RAT} 
    & Neutral    & 16 & 7 \\
    & CoT        & 16 & 1 \\
    & Anti-CoT   & 16 & 10 \\
\midrule
\multirow{3}{*}{BIG-Bench} 
    & Neutral    & 15 & 1 \\
    & CoT        & 15 & 1 \\
    & Anti-CoT   & 15 & 1 \\
\bottomrule
\end{tabular}
\caption{Steering intervention hyperparameters used for each dataset and prompt condition.}
\label{tab:steering_hparams}
\end{table*}

\section*{Computational Setup}

All experiments were conducted on a high-performance computing node equipped with 8 NVIDIA H100 GPUs (each with 80GB VRAM). We used PyTorch 2.1 and Hugging Face’s Transformers library for model loading and inference. Steering interventions were implemented through forward hooks on hidden states during inference. Each evaluation run was parallelized across GPUs using batch-level distribution. Prototype clustering and projection computations were performed on the same hardware, with minimal additional overhead relative to inference. No fine-tuning or model weight modifications were performed. All results reflect inference-time interventions on the pretrained LLaMA-3-Instruct-8B model.
\section*{Prompt-Length and Token Statistics}
To evaluate the economic efficiency of different steering strategies, we collect output token statistics across datasets and prompting conditions. For each combination of dataset, prompt type, and method (No Steering, DoM, PDS), we report the average number of tokens generated and the standard deviation.
Across all datasets, Prototype-Based Dynamic Steering (PDS) consistently produces shorter or comparable outputs to DoM, especially under CoT and Neutral prompting, while maintaining or improving accuracy. Under Anti-CoT prompts, all methods yield similarly minimal output, as expected by design. Detailed statistics are presented in Table~\ref{tab:token_stats}.

\section*{Licenses and Usage}
All models and datasets used in this study are publicly available and used in accordance with their respective licenses. Experiments involving LLaMA-3 models were conducted under their original open-source licenses (Llama 3 Community License Agreement). Benchmark datasets including GSM8K, AQuA-RAT, and BIG-Bench are licensed for research use (MIT or and Apache 2.0, respectively), and we adhere to all associated terms.

\begin{table*}[t]
\centering
\small
\begin{tabular}{clcccccc}
\toprule
\multirow{2}{*}{\textbf{Dataset}} & \multirow{2}{*}{\textbf{Prompt Type}}
& \multicolumn{2}{c}{\textbf{No Steering}} 
& \multicolumn{2}{c}{\textbf{DoM}} 
& \multicolumn{2}{c}{\textbf{PDS}} \\
\cmidrule(lr){3-4} \cmidrule(lr){5-6} \cmidrule(lr){7-8}
& & Avg. & Std. & Avg. & Std. & Avg. & Std. \\
\midrule
\multirow{3}{*}{\textbf{AQuA-RAT}} & Neutral     & 287.31 & 492.82 & 256.96 & 358.54 & \textbf{254.58} & \textbf{357.60} \\
 & CoT         & 275.61 & 358.16 & 303.09 & 440.70 & \textbf{258.03} & \textbf{146.45} \\
 & Anti-CoT    & 10.05  & 1.70   & 10.05  & 1.69   & 10.05  & 1.70 \\
\midrule
\multirow{3}{*}{\textbf{GSM8k}} & Neutral  & 155.12    & 61.56    & 158.74    & 55.69    & \textbf{126.32}    & \textbf{132.32} \\
 & CoT      &  182   & 121.86    & 180.27    & 58.92    & \textbf{90.29}    & \textbf{100.51} \\
 & Anti-CoT & 7.03    & 3.54    & 6.93    & 2.91    & \textbf{6.39}    & \textbf{4.25} \\
\bottomrule
\end{tabular}
\caption{Token length statistics (average and standard deviation) across datasets and prompt types for each steering method.}
\label{tab:token_stats}
\end{table*}





\end{document}